\documentclass[lettersize,journal]{IEEEtran}
\usepackage{amsmath,amsfonts}
\usepackage{algorithmic}
\usepackage{algorithm}
\usepackage{array}
\usepackage[caption=false,font=normalsize,labelfont=sf,textfont=sf]{subfig}
\usepackage{textcomp}
\usepackage{stfloats}
\usepackage{url}
\usepackage{verbatim}
\usepackage{graphicx}
\usepackage{cite}

\usepackage{xcolor}
\usepackage{booktabs}
\usepackage{multirow}
\usepackage{pifont}
\usepackage{colortbl}
\usepackage{gensymb}
\usepackage{tabularx}
\usepackage{makecell}
\usepackage{subfig}
\usepackage{orcidlink}

\definecolor{mygreen}{RGB}{218,235,192}
\definecolor{editpurple}{RGB}{128,0,170}
\definecolor{brown}{RGB}{165,42,42} 

\begin{document}

\title{Learning to Anchor Visual Odometry: KAN-Based Pose Regression for Planetary Landing}

\author{Xubo Luo~\orcidlink{0000-0002-7388-5749},
        Zhaojin Li~\orcidlink{0000-0002-8209-5749},
        Xue Wan~\orcidlink{0000-0002-0659-2559},
        Wei Zhang~\orcidlink{0000-0003-4088-566X},
        Leizheng Shu~\orcidlink{0000-0002-7287-1142}
\thanks{Received September 23, 2025; Accepted December 30, 2025.}
\thanks{Xubo Luo is with the University of Chinese Academy of Sciences, Beijing 100101, China (e-mail: luoxubo23@ucas.ac.cn).}
\thanks{Zhaojin Li, Xue Wan, Wei Zhang and Leizheng Shu are with the Technology and Engineering Center for Space Utilization, Chinese Academy of Sciences, Beijing 100094, China (email: lizhaojin, wanxue, zhangwei, shuleizheng@csu.ac.cn).}
}

\markboth{Journal of \LaTeX\ Class Files,~Vol.~14, No.~8, August~2021}%
{Shell \MakeLowercase{\textit{et al.}}: A Sample Article Using IEEEtran.cls for IEEE Journals}


\maketitle

\begin{abstract}
Accurate and real-time 6-DoF localization is mission-critical for autonomous lunar landing, yet existing approaches remain limited: visual odometry (VO) drifts unboundedly, while map-based absolute localization fails in texture-sparse or low-light terrain. We introduce KANLoc, a monocular localization framework that tightly couples VO with a lightweight but robust absolute pose regressor. At its core is a Kolmogorov–Arnold Network (KAN) that learns the complex mapping from image features to map coordinates, producing sparse but highly reliable global pose anchors. These anchors are fused into a bundle adjustment framework, effectively canceling drift while retaining local motion precision. KANLoc delivers three key advances: (i) a KAN-based pose regressor that achieves high accuracy with remarkable parameter efficiency, (ii) a hybrid VO–absolute localization scheme that yields globally consistent real-time trajectories ($\geq$15 FPS), and (iii) a tailored data augmentation strategy that improves robustness to sensor occlusion. On both realistic synthetic and real lunar landing datasets, KANLoc reduces average translation and rotation error by 32\% and 45\%, respectively, with per-trajectory gains of up to 45\%/48\%, outperforming strong baselines.
\end{abstract}


\section{Introduction}
Lunar landing missions demand precise 6-DoF pose estimation, where even minor inaccuracies can jeopardize mission success~\cite{prause2024fatal, baker2024comprehensive}. Traditional navigation approaches often rely on high-power LiDAR systems~\cite{amzajerdian2022analysis} or complex IMU-based sensor fusion~\cite{andolfi2024moon}, which impose significant constraints on payload, power, and complexity, all critical factors for resource-constrained missions. Vision-based localization offers a lightweight yet powerful alternative, providing rich environmental information with minimal hardware overhead.

However, the lunar environment presents unique challenges. Relative visual localization methods, such as Visual Odometry (VO), are locally stable but inevitably accumulate drift over long trajectories~\cite{fabian2014error}. Absolute Visual Localization (AVL), which aligns the lander's camera view with a prior satellite map, can correct this drift. Yet, traditional AVL pipelines, based on feature matching and Perspective-n-Point (PnP) solvers~\cite{ostrogovich2024dual}, show insufficient robustness under lunar conditions. Available digital elevation models (DEMs) are often too coarse ($\approx$10 m resolution) for low-altitude navigation~\cite{huang2025fusion}, while the repetitive textures and harsh illumination of the lunar surface lead to sparse and ambiguous feature correspondences. This leaves a critical gap: VO provides continuity but drifts, whereas AVL offers global accuracy but is often unreliable in practice.

\begin{figure}[thp]
    \centering
    \includegraphics[width=\linewidth]{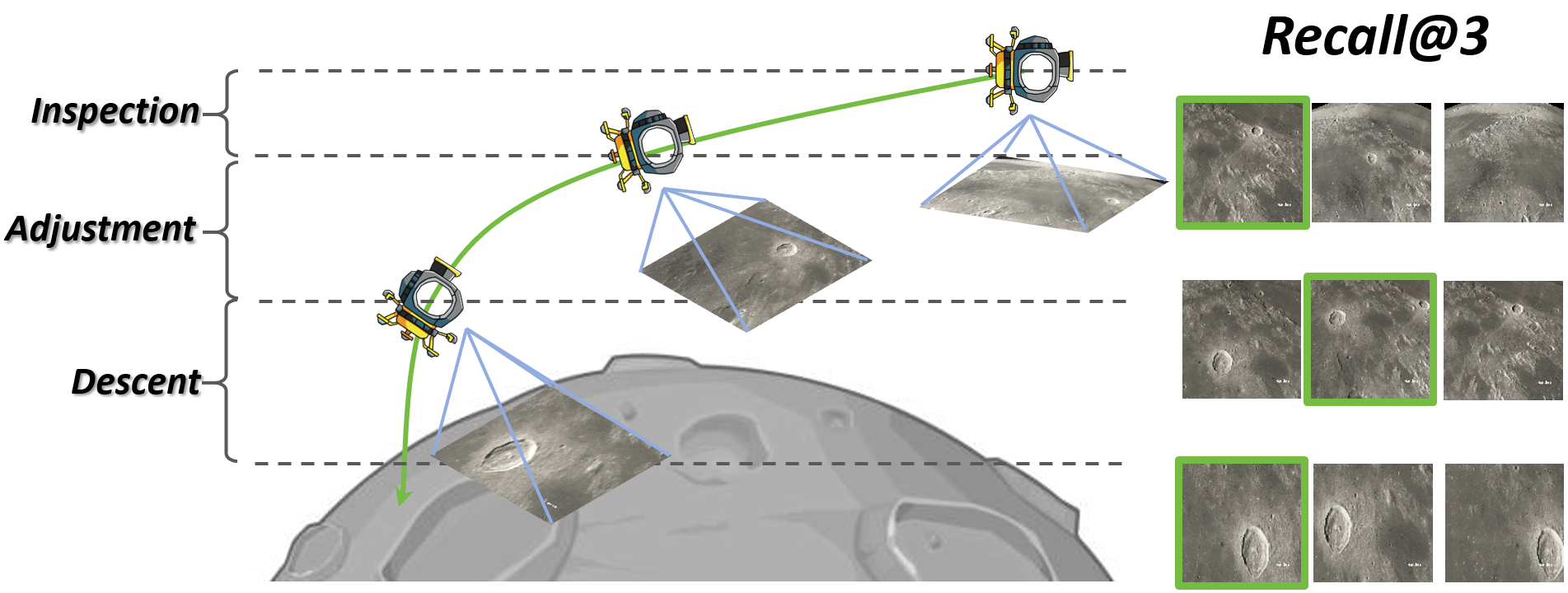}
    \caption{Illustration of lunar landing phases, such as those in the Chang'e-3 mission, which require continuous and accurate localization across different altitudes and sensor conditions.}
    \label{fig:fig1}
\end{figure}

To bridge this gap, we introduce KANLoc, a framework that systematically integrates high-frequency VO with low-frequency absolute pose corrections from a novel AVL module. The key innovation is an absolute pose estimator based on Kolmogorov-Arnold Networks (KANs). While conventional deep learning models can regress pose, they often require large datasets that are scarce for lunar scenarios and can be too computationally heavy for onboard deployment. In contrast, KANs replace the fixed, scalar weights of standard Multilayer Perceptrons (MLPs) with learnable univariate functions, decomposing the high-dimensional image-to-map relationship into a series of smooth, localized transformations. This structure provides superior parameter and sample efficiency, enabling robust image-map alignment from limited data while remaining lightweight. These properties make KANs particularly well-suited to the cross-domain regression problem between lander imagery and satellite maps. Within KANLoc, our KAN-based estimator produces sparse but accurate absolute pose anchors, which are fused with a continuous VO tracker through a tightly-coupled bundle adjustment. This fusion preserves local motion smoothness while enforcing global consistency. The contributions of this paper are threefold:

\begin{itemize}
\item We propose a KAN-based absolute pose regressor that performs robust end-to-end estimation, overcoming the limitations of feature-based methods in texture-sparse lunar environments.
\item We introduce a tightly-coupled hybrid framework that fuses high-frequency VO with sparse absolute KAN anchors, achieving globally consistent localization with high efficiency.
\item We develop a mask augmentation strategy to bridge the Sim-to-Real gap, enabling the model to generalize to real flight data despite severe self-occlusions.
\end{itemize}

\section{Related Work}

\subsection{Monocular Localization}
Monocular localization infers 6-DoF pose via Relative (RVL) and Absolute (AVL) Visual Localization.

\textbf{RVL} tracks motion between successive frames. Examples include the Nova-C lander~\cite{molina2022visual} and Kalman-based methods~\cite{uzun2024dual}. Recent neural approaches address the limitations of handcrafted features by enforcing metric scale via geometric constraints~\cite{zhao2021deep} or patch-based correspondences~\cite{teed2023deep}. Despite providing real-time estimates, all VO methods suffer from cumulative drift.

\textbf{AVL} corrects drift by aligning images to a map. Classic methods match features to solve PnP problems~\cite{tong2024illumination}; {in the context of planetary landing, Christian et al.~\cite{christian2021image} provide a comprehensive review of such Terrain Relative Navigation (TRN) techniques.} However, matching speed and resolution disparities often limit performance~\cite{liu2022generative}. {While probabilistic feature alignment methods like PixLoc~\cite{sarlin2021featurelearningrobustcamera} offer improved robustness, they can be computationally intensive during inference.} Consequently, end-to-end learning frameworks have emerged, categorized into direct pose regressors and scene coordinate regressors. While powerful, they struggle with generalization and computational cost~\cite{dong2024reloc3r, zheng2025scene}, with diffusion-based refinements~\cite{yang2023diffusion} adding further overhead. Our work leverages the novel KAN architecture for direct pose regression, balancing high accuracy with efficiency.

\subsection{Hybrid Localization}
Hybrid localization fuses relative and absolute measurements to prevent drift, typically via Bundle Adjustment (BA)~\cite{chang2019gnss,holmes2023efficient}. We adopt a tightly-coupled approach, integrating absolute priors directly as optimization factors for superior robustness compared to loosely-coupled methods. While early works used GNSS~\cite{huang2021camera}, its latency is prohibitive for cislunar space, prompting a shift to visual map matching~\cite{wan2022terrain,luo2024jointloc}. Crucially, unlike existing methods that rely solely on position priors, our framework incorporates full 6-DoF absolute pose constraints.

\subsection{Kolmogorov-Arnold Networks (KAN)}
Inspired by the Kolmogorov-Arnold theorem, KANs replace fixed nodal activations with learnable univariate splines on edges. This architecture enables efficient modeling of complex functions with fewer parameters than standard MLPs. We leverage this to map high-dimensional DINO-v2 features to the $\mathbb{SE}(3)$ pose manifold, as KANs can capture geometric structure more effectively than monolithic Transformers or CNNs. While promising in other domains~\cite{somvanshi2024survey}, this is the first application of KANs to absolute pose regression, yielding a lightweight and robust estimator for robotic localization.

\begin{figure*}[htp]
    \centering
    \includegraphics[width=0.98\linewidth]{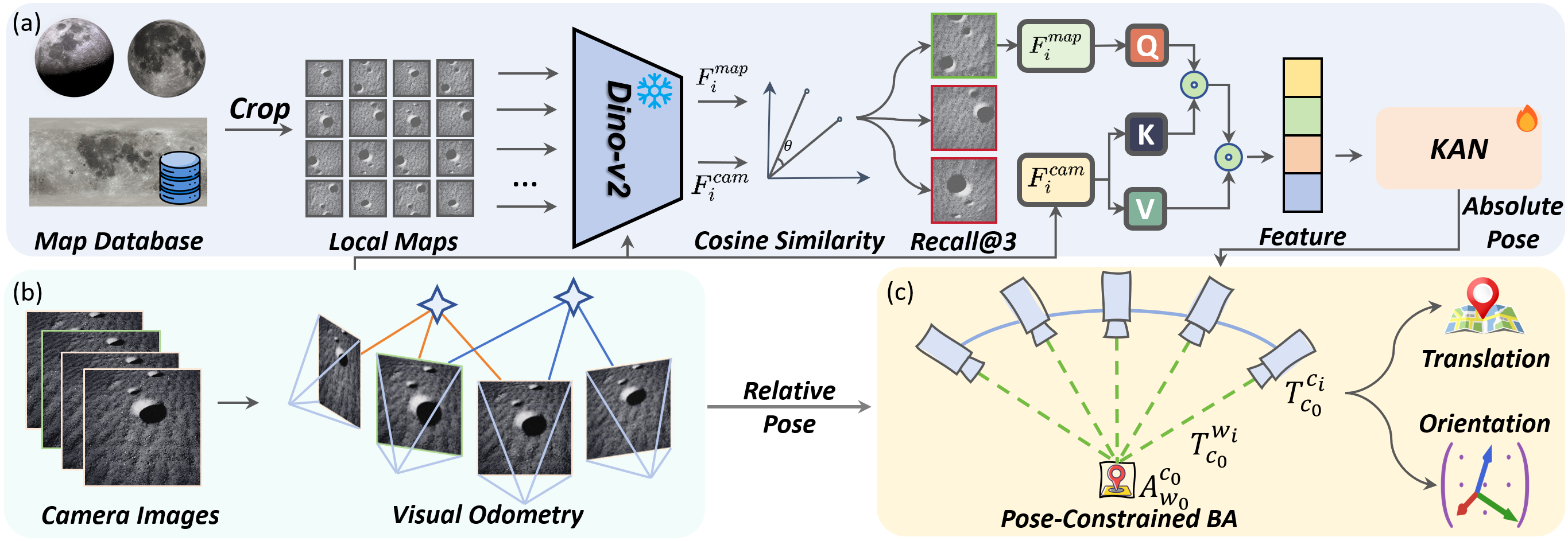}
    \caption{The proposed KANLoc framework integrates: (a) AVL, combining coarse retrieval with KAN-based 6-DoF refinement; (b) RVL, utilizing standard visual odometry for high-frequency tracking; and (c) a bundle adjustment back-end ensuring global consistency via AVL constraints.}
    \label{fig:pipeline}
\end{figure*}

\section{Methodology}
\label{sec:method}

Given a sequence of images $(I^i)_{i=1}^N$ from the lander and prior reference maps, our goal is to estimate the lander's real-time 6-DoF pose. We explicitly state that pose transformations are composed from right-to-left, where $T^a_b$ denotes the pose of frame {$a$ in frame $b$}. As shown in Fig.~\ref{fig:pipeline}, our framework combines high-frequency visual odometry for local motion tracking with low-frequency absolute pose estimation from our novel KAN-based regressor. These outputs are fused via absolute-pose-constrained bundle adjustment for a globally consistent trajectory. This design explicitly targets two key challenges in autonomous landing: (i) drift accumulation in relative pose estimates and (ii) global alignment under scale ambiguity.

\subsection{KAN-based Absolute Pose Regressor}
We propose an AVL module that estimates the 6-DoF pose of the lander relative to a global map. In contrast to conventional MLP or transformer-based regressors, our design leverages the KAN, which offers provable universal approximation via the Kolmogorov–Arnold representation theorem, improved interpretability, and strong empirical efficiency for real-time inference.

\textbf{Coarse Localization via Image Retrieval.} {We retrieve the top-$K$ ($K=5$) map tiles (2.0 m/px, 20\% overlap in $w_0$) using DINO-v2~\cite{oquab2023dinov2} cosine similarity. The search is constrained to a local neighborhood ($W=9$) around the previous match for efficiency.} To mitigate tracking failures, if the similarity score falls below a threshold $\tau$, the system reverts to global search in the next frame.

\textbf{Pose Regression with KAN.} Given the DINO descriptors $\mathbf{f}^i_c \in \mathbb{R}^n$ (camera) and $\mathbf{f}^i_m \in \mathbb{R}^n$ (retrieved map), we fuse them via a lightweight learned gate:
\begin{equation}
    \alpha = \sigma\!\left(\frac{(W_q \mathbf{f}^i_c)^\top (W_k \mathbf{f}^i_m)}{\sqrt{d}}\right),
\end{equation}
\begin{equation}
    \mathbf{x}^i = \alpha\, (W_v \mathbf{f}^i_m) + (1-\alpha)\, (W_c \mathbf{f}^i_c),
\end{equation}
where $W_q,W_k,W_v,W_c$ are learnable projections and $\sigma(\cdot)$ is the sigmoid. This gating lets the camera descriptor modulate the contribution of the retrieved map descriptor. We also evaluate a token-level cross-attention variant in the supplement.

The fused representation $\mathbf{x}^i\in\mathbb{R}^n$ is then processed by our KAN regressor. Unlike conventional MLP layers with fixed nonlinearities, KAN layers parameterize nonlinear transformations using learnable spline bases, providing adaptive capacity to capture high-dimensional variations in planetary imagery. Specifically, the inner and outer functions are modeled as linear combinations of parametric and B-spline basis functions:
\begin{equation}
    \psi(x) = w_b\,\frac{x}{1 + e^{-x}} + w_s\,\sum_i c_i B_i(x),
\end{equation}
where $B_i(x)$ denotes the $i$-th B-spline basis, $w_b$ and $w_s$ are learnable weights, and $c_i$ are coefficients. To alleviate the GPU overhead of recursive B-spline computation, we adopt the Reflectional Switch Activation Function (RSWAF)~\cite{ta2024bsrbf}, which provides an efficient closed-form approximation:
\begin{equation}
    B_i(x) \approx 1 - \left(\tanh\left(\frac{x - x_i}{h}\right)\right)^2.
\end{equation}

Our KAN regressor comprises three layers with widths and a spline grid size of 5. At inference, {the model achieves up to 20 Hz on a RTX 4090 GPU}. After {$L=3$ layers}, a final linear head regresses the 6-DoF pose $(\hat{R}^i, \hat{t}^i)$. To ensure $\hat{R}^i\in\mathrm{SO}(3)$, we predict a 6D rotation~\cite{zhou2019continuity}, orthonormalizing it via Gram–Schmidt and negating the last column if $\det(\hat{R}^i) < 0$. We supervise the network using:
\begin{equation}
    \mathcal{L}_{\mathrm{pose}} = \arccos\left(\mathcal{C}\!\left(\frac{\mathrm{tr}(\hat{R}^i {R^i_{\mathrm{gt}}}^\top) - 1}{2}\right)\right) + \lambda \|\hat{t}^i - t^i_{\mathrm{gt}}\|_2^2,
\end{equation}
{where $\mathcal{C}(\cdot)$ clamps inputs to} $[-1+\epsilon, 1-\epsilon]$ ($\epsilon=10^{-7}$) {for numerical stability, and $\lambda=10$ balances the translation and rotation terms. The estimated absolute pose is denoted $T^{c_i}_{w_0} \in \mathbb{SE}(3)$.}

\begin{algorithm}[t]
\small 
\caption{KANLoc AVL Pipeline}
\label{alg:kan_avl_compact}
\begin{algorithmic}[1] 
\STATE \textbf{In:} Img $I_c^i$, Map $\mathcal{D}$, $j_{\text{prev}}$, $W$, $K$; \textbf{Out:} Pose $T^{c_i}_{w_0}$, Info $\Omega_i$
\STATE Get Top-$K$ candidates $\{m_k\}$ in window $W$ via DINO sim: ${\mathbf{f}_c^i}^\top \mathbf{f}_{m_k}$
\FOR{each candidate $m_k$}
  \STATE Regress pose: $(\hat{R}, \hat{\mathbf{t}}) \leftarrow \mathrm{Ortho}(\mathrm{KAN}(\mathrm{Fuse}(\mathbf{f}_c^i, \mathbf{f}_{m_k})))$
  \STATE Assemble $T_k^{c_i}{_{w_0}}$ and compute score $s_k \!\leftarrow\! \alpha\,\cos(\mathbf{f}_c^i,\mathbf{f}_{m_k}) + (1-\alpha)\,\mathrm{conf}(T_k^{c_i}{_{w_0}})$
\ENDFOR
\STATE Select $k^* \leftarrow \arg\max_k s_k$ and set $T^{c_i}{_{w_0}} \leftarrow T_{k^*}^{c_i}{_{w_0}}$
\STATE Calc residual $\mathbf{e}_i$; \textbf{return} $(\mathbf{e}_i^\top \Sigma^{-1} \mathbf{e}_i < \chi^2_\tau)$ ? $(T^{c_i}{_{w_0}}, \Sigma^{-1})$ : \textbf{Reject}
\end{algorithmic}
\end{algorithm}
The scoring function in Algorithm~\ref{alg:kan_avl_compact} combines the retrieval similarity with a pose confidence score, where $\alpha$ is a weighting hyperparameter {(empirically set to $\alpha=0.7$)}. 
{The confidence $\mathrm{conf}(\cdot)$ is set to a constant as the current KAN regressor is deterministic. Therefore, candidate scoring relies on retrieval similarity combined with the subsequent chi-squared gating, which we found sufficient for robust outlier rejection.}

\subsection{Relative Pose Estimation}
For high-frequency motion estimation, we employ a custom ORB-based monocular VO pipeline implemented in OpenCV~\cite{campos2021orb}. 
{We selected this CPU-efficient method to establish a heterogeneous computing architecture, reserving GPU resources solely for the KAN network to avoid hardware contention. Furthermore, unlike the patch-based DINO features used for global matching, ORB provides the efficient pixel-level geometric precision required for high-frequency tracking.}
Sparse keypoints are detected and described using ORB, and correspondences are established via a K-nearest neighbors matcher. To improve robustness, matches are filtered by the ratio test and geometric verification with RANSAC.

Once an initial motion estimate is obtained, 3D landmarks are triangulated. Camera poses are refined by minimizing the total reprojection error:
\begin{equation}
    E_{\mathrm{rep}} = \sum_{k} \bigl\|\mathbf{p}_k - \pi\bigl(\mathbf{K}(\mathbf{R}\mathbf{X}_k + \mathbf{t})\bigr)\bigr\|^2,
\end{equation}
where $\pi(\cdot)$ is the perspective projection function and $\mathbf{K}$ is the intrinsic matrix. The estimated relative pose is denoted as $T^{c_i}_{c_0} \in \mathbb{SE}(3)$. 

\subsection{Globally-Consistent Trajectory Estimation via Constrained BA}
The relative pose from VO is scale-ambiguous and defined in a local frame initialized at the start of motion. To correct drift and align the trajectory to the global frame, we fuse the VO estimates with the sparse absolute pose measurements from our KAN module. Unlike standard BA, our approach incorporates these absolute poses as priors in a pose-graph optimization. A similarity transformation $\mathcal{A}^{c_0}_{w_0}\in \mathrm{Sim}(3)$ is jointly optimized to align the relative VO frame to the global world frame:
\begin{equation}
    \hat{T}^{c_i}_{w_0} = \mathcal{A}^{c_0}_{w_0}T^{c_i}_{c_0},
\end{equation}
where $\hat{T}^{c_i}_{w_0}$ is the aligned $i$-th camera pose. The residual for an absolute pose measurement $T^{c_i}_{w_0}$ is defined on the manifold $\mathfrak{se}(3)$:
\begin{equation}
\begin{aligned}
    \mathbf{e}_i = \log\left((T^{c_i}_{w_0})^{-1} \hat{T}^{c_i}_{w_0}\right)^\vee 
     = \log\left((T^{c_i}_{w_0})^{-1} \mathcal{A}^{c_0}_{w_0} T^{c_i}_{c_0}\right)^{\vee}
\end{aligned}
\end{equation}
where $(\cdot)^\vee$ maps an element of the Lie algebra $\mathfrak{se}(3)$ to its vector representation. We construct a factor graph where camera poses are nodes, and edges represent either relative motion constraints from VO or absolute pose constraints from our KAN module. We employ g2o~\cite{kummerle2011g2o} to solve the full nonlinear least-squares problem, which jointly minimizes reprojection errors and absolute pose residuals: All factors are wrapped with a Huber kernel, and suspected outliers are first filtered using a chi-square gate.
\begin{equation}
    \mathcal{A}^* = \arg\min_{\mathcal{A}} \sum_{i \in \mathcal{K}} \mathbf{e}_i^\top \mathbf{\Omega}_i \mathbf{e}_i + \sum_{j \in \mathcal{M}} E_{\mathrm{rep}, j},
\end{equation}
where $\mathcal{K}$ is the set of keyframes with absolute pose measurements from the KAN, $\mathcal{M}$ is the set of all 3D map points, and $\mathbf{\Omega}_i$ is the information matrix for the absolute measurement. 
{In the measurement selection step (Algorithm 1), the confidence term is set to a constant because the current KAN regressor is deterministic; thus, candidate weighting relies primarily on retrieval similarity.}
The information matrix $\mathbf{\Omega}_i$ provides a principled way to weight the influence of each absolute measurement. The empirical covariance $\Sigma$ is estimated on a held-out validation set by aggregating pose regression errors, and its inverse $\mathbf{\Omega}_i=\Sigma^{-1}$ is frozen during testing. 
To filter dynamic outliers, we set the chi-square gating threshold at {$\chi^2_\tau \approx 16.81$}, corresponding to a p-value of 0.99 with 6 degrees of freedom {(covering both translation and rotation)}. This process fuses absolute and relative poses, yielding a smooth and globally consistent trajectory.

\section{Experiments}
In this section, we first perform experiments on a synthetic dataset to investigate the influence of lighting and terrain on the localization accuracy of our method. Following this, we evaluate on the real-world landing sequence of Chang'e-3 to assess the generalization ability of our method.

\subsection{Experiments on Synthetic Datasets}
Visual localization accuracy against a prior map is heavily influenced by factors like lighting and terrain. In this section, we systematically investigate these effects using synthetic datasets.

\begin{figure}[htp]
        \centering
        \subfloat[Chang'e-3]{
            \includegraphics[width=0.4\linewidth, height=0.25\linewidth]{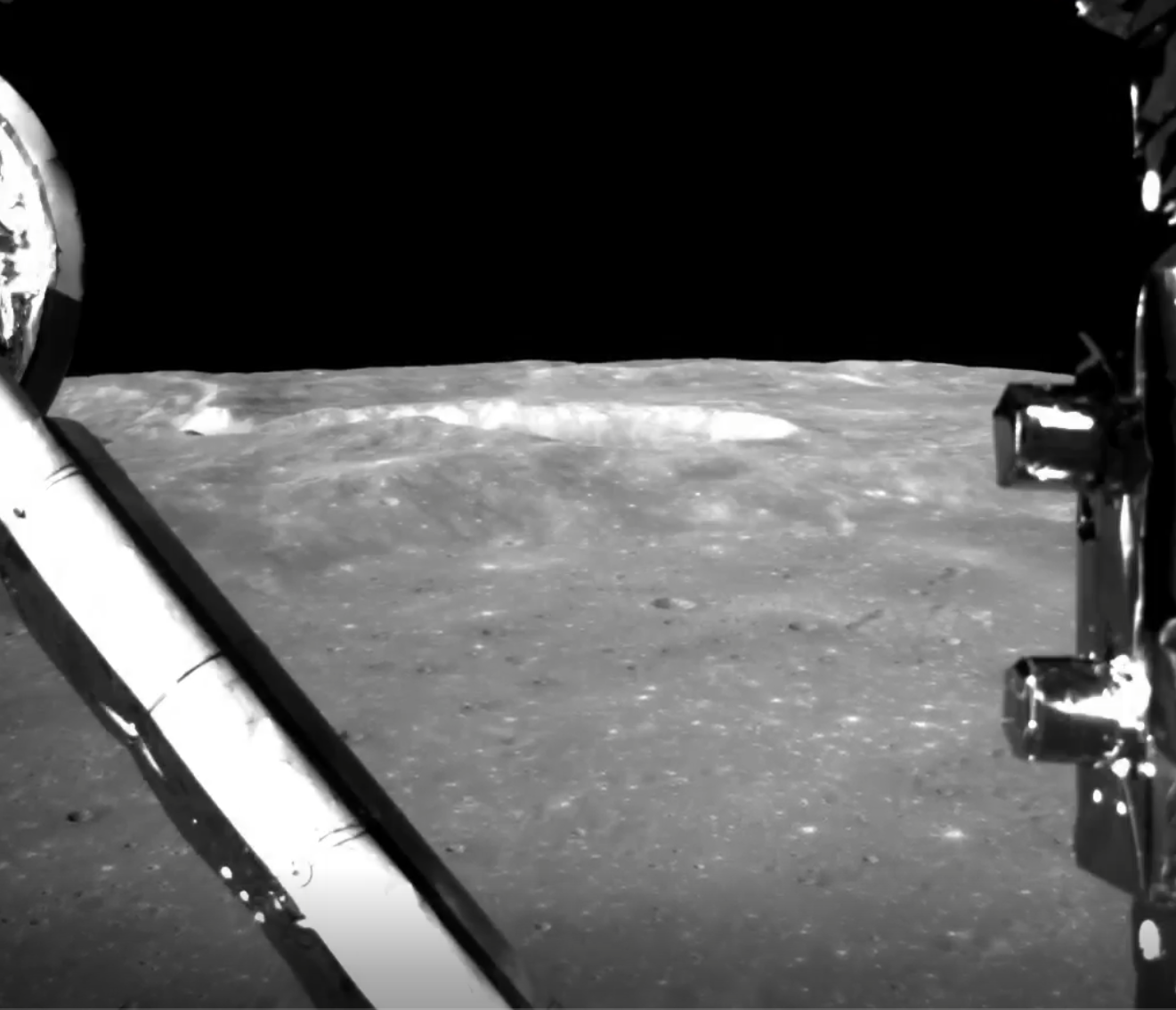}
            \label{fig:sim_sample}
        }
        \subfloat[Simulated Lunar]{
            \includegraphics[width=0.4\linewidth, height=0.25\linewidth]{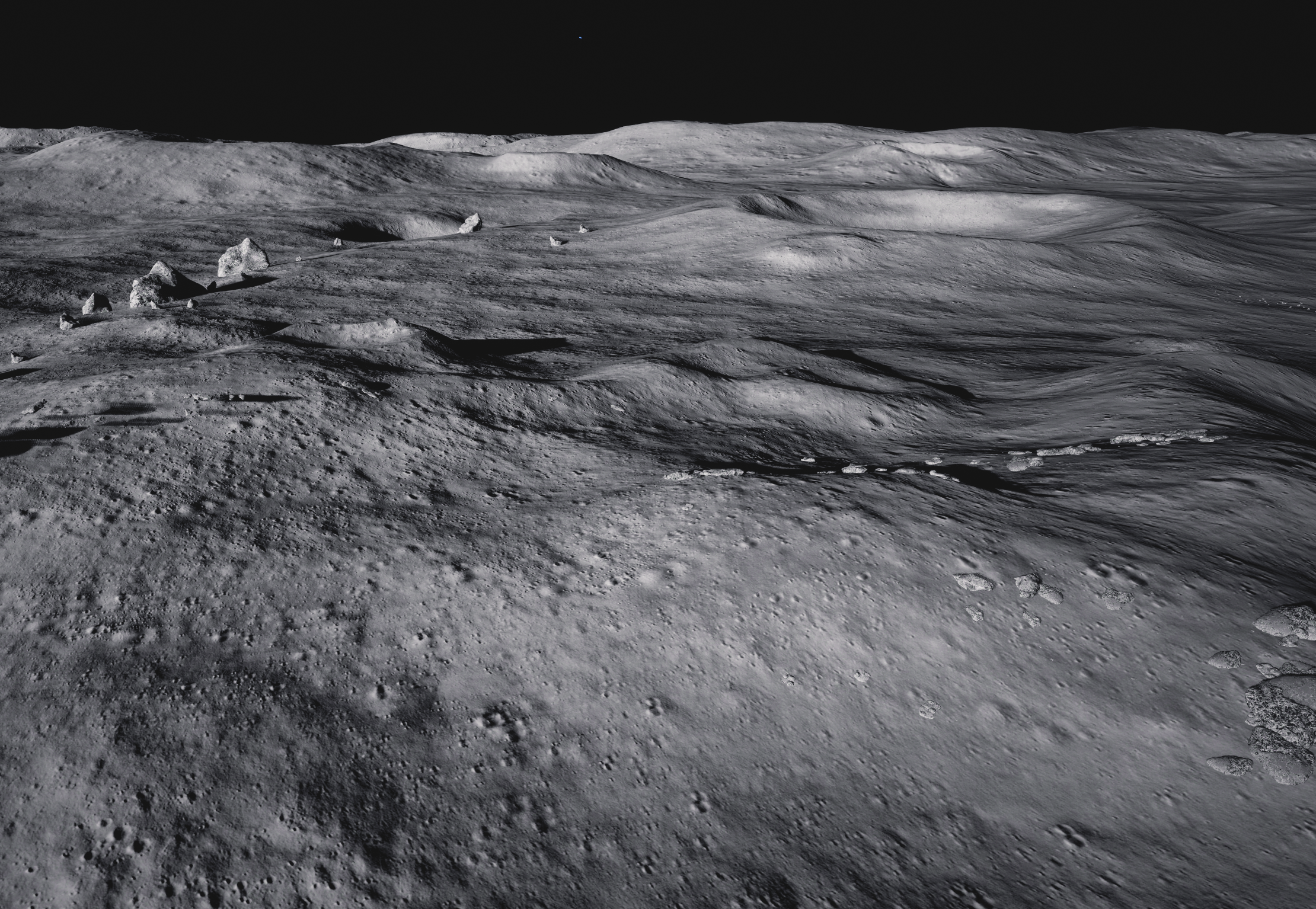}
            \label{fig:real_sample}
        }
        \caption{Sample images from (a) real Chang'e-3 imagery and (b) high-fidelity simulation used for training or evaluation.}
        \label{fig:dataset_samples}
    \end{figure}

\subsubsection{Synthetic Dataset Generation}
We build a synthetic lunar landing dataset using Unreal Engine and AirSim~\cite{shah2018airsim}, simulating the Chang'e-3 landing site. The dataset includes four trajectories with lengths ranging from 1800m to 2000m, covering varying illumination (0.2-1.0) and slope angles (0-20\degree). The lander's trajectory simulates near-ground inspection (1000m), angle adjustment (500-1000m), and final descent (0-500m) stages\footnote{We acknowledge that real missions often face dust-induced visual degradation at low altitudes ($<$10m), typically necessitating a switch to non-visual sensors.}.

\begin{table*}[htp] 
        \caption{{RMSE of the proposed method and baselines on the synthetic dataset. $^{\dagger}$ indicates methods evaluated on CPU; others are evaluated on a GPU.}}
        \label{tab:synthetic_results}
        \centering
        
        \resizebox{\textwidth}{!}{ 
            \begin{tabular}{llcccccccccccc}
            \toprule
            \multicolumn{2}{c}{\multirow{2}{*}{\textbf{Method}}} &
            \multicolumn{5}{c}{\textbf{Err. Trans. (m)$\downarrow$}} &
            \multicolumn{5}{c}{\textbf{Err. Rot. (\degree)$\downarrow$}} &
            \multirow{2}{*}{\textbf{FPS}$\uparrow$} &
            \multirow{2}{*}{\textbf{Params (M)$\downarrow$}} \\
            \cmidrule(lr){3-7} \cmidrule(lr){8-12}
            & & T1 & T2 & T3 & T4 & Avg. & T1 & T2 & T3 & T4 & Avg. & & \\
            \midrule
            \multirow{5}{*}{\textbf{RVL}} 
                & ORB-SLAM2$^{\dagger}$   & 16.85 & 16.12 & 15.90 & 15.45 & 16.08 & 18.92 & 18.10 & 17.85 & 17.30 & 18.04 & 19.20 & -- \\
                & ORB-SLAM3$^{\dagger}$   & 10.65 & 11.02 & 11.35 & 11.10 & 11.03 & 13.80 & 14.12 & 14.45 & 13.98 & 14.09 & \textbf{20.14} & -- \\
                & DPVO        & 14.10 & 14.35 & 14.60 & 14.85 & 14.48 & 15.45 & 15.70 & 15.95 & 16.20 & 15.83 & 14.80 & \textbf{2.3} \\
                & NeRF-VO     & 13.98 & 14.22 & 14.45 & 14.68 & 14.33 & 15.32 & 15.55 & 15.78 & 16.01 & 15.67 & 14.50 & 131.2 \\
                & LEAP-VO     & 13.85 & 14.08 & 14.30 & 14.52 & 14.19 & 15.19 & 15.41 & 15.63 & 15.85 & 15.52 & 14.90 & 43.7 \\
            \midrule
            \multirow{5}{*}{\textbf{AVL}} 
                & HLoc     & 19.82 & 22.05 & 24.76 & 27.49 & 23.53 & 22.66 & 25.38 & 28.10 & 30.82 & 26.74 & 9.65  & 73.6 \\
                & Reloc3r     & 25.78 & 30.01 & 35.23 & 40.45 & 32.87 & 30.12 & 35.34 & 40.56 & 45.78 & 37.95 & 4.39  & 103.5 \\
                & RelPose++   & 30.60 & 35.83 & 40.05 & 45.27 & 37.94 & 40.94 & 45.16 & 50.38 & 55.60 & 48.02 & 3.16  & 28.4 \\
                & Direct-PoseNet & 15.72 & 20.95 & 25.17 & 30.39 & 23.06 & 25.06 & 30.28 & 35.50 & 40.72 & 32.89 & 6.68 & 22.6 \\
                & PoseDiffusion & 20.65 & 25.88 & 30.10 & 35.32 & 27.99 & 30.99 & 35.21 & 40.43 & 45.65 & 38.07 & 4.01 & 41.2 \\
            \midrule
            \multirow{1}{*}{\textbf{Hybrid}}
                & \textbf{KANLoc (Ours)} 
                & \textbf{5.90} & \textbf{6.05} & \textbf{8.10} & \textbf{7.95} & \textbf{7.50} 
                & \textbf{7.65} & \textbf{7.30} & \textbf{8.15} & \textbf{8.00} & \textbf{7.78} 
                & 16.35 & 35.6 \\
            \bottomrule
            \end{tabular}
        } 
    \end{table*}

\subsubsection{Implementation Details}
We compare against representative monocular localization methods, including RVL (ORB-SLAM2~\cite{mur2017orb}, ORB-SLAM3~\cite{campos2021orb}, DPVO~\cite{teed2023deep}, NeRF-VO~\cite{naumann2024nerf}, LEAP-VO~\cite{chen2024leap}) and AVL (HLoc~\cite{sarlin2019coarse}, Reloc3r~\cite{dong2024reloc3r}, RelPose++~\cite{lin2024relpose++}, Direct-PoseNet~\cite{chen2021direct}, PoseDiffusion~\cite{wang2023posediffusion}). To ensure a fair and rigorous comparison, all learning-based baselines were re-trained from scratch on our synthetic lunar dataset using their publicly available implementations. We adopted a unified training protocol for all methods: Adam optimizer, initial learning rate $1\times10^{-3}$ with $\times0.1$ step decay every 20 epochs (60 total), batch size 32, weight decay $10^{-4}$, and input resolution of $256\times256$. The loss weight $\lambda$ for pose regression methods was set to 10. This protocol ensures that performance differences can be attributed to model architecture and methodology rather than disparities in training data or optimization. 
Unless otherwise stated, performance is measured on a single NVIDIA RTX 4090. {We employ a dual-resolution strategy: $256\times256$ for KAN inference to ensure speed, and $512\times512$ for ORB tracking to maintain geometric precision. VO threading is fixed, AVL fires every 10 frames, and retrieval uses GPU-based descriptors.}

\subsubsection{Results and Analysis}
Table~\ref{tab:synthetic_results} shows the quantitative results. Among RVL methods, ORB-SLAM3 is the most accurate but still exhibits significant drift, with an average translation error of 11.03m. AVL methods generally show much higher errors and slower speeds, highlighting the difficulty of direct regression even when trained on the target domain.

Our proposed KANLoc framework consistently achieves the lowest error across all trajectories, registering an average translation error of 7.50m and a rotation error of 7.78\degree. This represents a significant improvement, reducing the translation and rotation errors of the next-best method, ORB-SLAM3, by 32.0\% and 44.8\%, respectively. Notably, KANLoc maintains robust performance under challenging conditions like low illumination and rough terrain (Trajectories 2 and 4), where other methods degrade. The significant performance drop of other AVL methods in low light (e.g., Reloc3r on Trajectory 4, with a 40.45m translation error) is likely due to the collapse of distinctive visual cues that their backbones rely on. {Similarly, the structure-based baseline HLoc performs well in distinctive terrains (T1, T2) but struggles in texture-poor regions (T3, T4), leading to higher overall errors compared to KANLoc.} In contrast, our approach benefits from the robustness of the pre-trained DINO features and the KAN's ability to learn a smooth mapping function that is less sensitive to illumination variance. The hybrid approach effectively mitigates drift while ensuring global consistency.

In terms of efficiency, KANLoc runs in real-time at 16.35 FPS, a speed that is significantly faster than all AVL methods and competitive with most RVL methods. This validates the parameter efficiency of KANs compared to larger models used in other methods, such as Reloc3r (103.5M) and NeRF-VO (131.2M). {Quantitatively, the proposed neighborhood search strategy significantly reduces computational load. Compared to the global search which scans 165 regions ($\approx 1650$ ms), our method processes only $\leq 9$ local patches ($\approx 90$ ms), achieving an $18\times$ speedup.}

\subsection{Experiments on Real Flight Dataset}
We now evaluate our method on real flight data from the Chang'e-3 mission to assess its generalization to real-world conditions.

\subsubsection{Dataset Preparation}
We processed the publicly available landing video of Chang'e-3 into an image sequence at 25 fps, resulting in 1,060 frames over a 1268.69-meter trajectory. This dataset is challenging due to occlusions from the lander's own components (e.g., legs, support structures), which can obscure visual features. To improve robustness, we applied mask augmentation to our synthetic training data, simulating these occlusions. This forces the model to learn from partial views, improving its generalization. Details on the augmentation strategy are in Section~\ref{sec:mask_ablation}. Because absolute ground truth is unavailable, we align predicted trajectories to a geo-registered reference using Sim(3) alignment~\cite{umeyama1991least}. We acknowledge that this evaluation protocol can mask certain global errors like scale drift but remains a standard and necessary practice for such datasets; we will release our alignment scripts to ensure reproducibility.

\subsubsection{Implementation Details}
We use the same set of competing methods and evaluation metrics (RMSE of translation and rotation, FPS) as in the synthetic experiments.

\subsubsection{Results and Analysis}
Table~\ref{tab:real_results} presents the results on the real flight dataset. Our proposed KANLoc method significantly improves over the selected baselines in accuracy while maintaining real-time performance. It achieves a translation error of 12.50m and a rotation error of 9.25\degree.

RVL methods like LEAP-VO perform reasonably well but still suffer from accumulated drift. {AVL methods show severely degraded performance on real data. Specifically, Reloc3r and RelPose++ exhibit extreme translational jitter and Z-axis drift rather than a complete loss of odometry. This failure stems from overfitting to synthetic textures, leading to scale misidentification on real imagery. }
This domain gap arises from subtle differences in sensor noise, motion blur, and the unique photometric properties of lunar regolith that are difficult to perfectly replicate in simulation.

KANLoc demonstrates remarkable robustness, achieving the lowest translation and rotation error. Compared to the best RVL method (LEAP-VO), our framework improves translation accuracy by 2.34\% and rotation accuracy by 23.6\%. This highlights the effectiveness of our drift correction mechanism. Crucially, it maintains a practical 15.52 FPS. {The scale ambiguity is a critical concern in monocular systems. Global ATE can often hide potentially dangerous errors along the approach axis. Table~\ref{tab:real_results} explicitly reports height error, where KANLoc achieves a significant 35.3\% improvement over LEAP-VO, demonstrating its ability to maintain scale consistency.}

\begin{figure*}[htp]
    \centering
    \includegraphics[width=.9\linewidth]{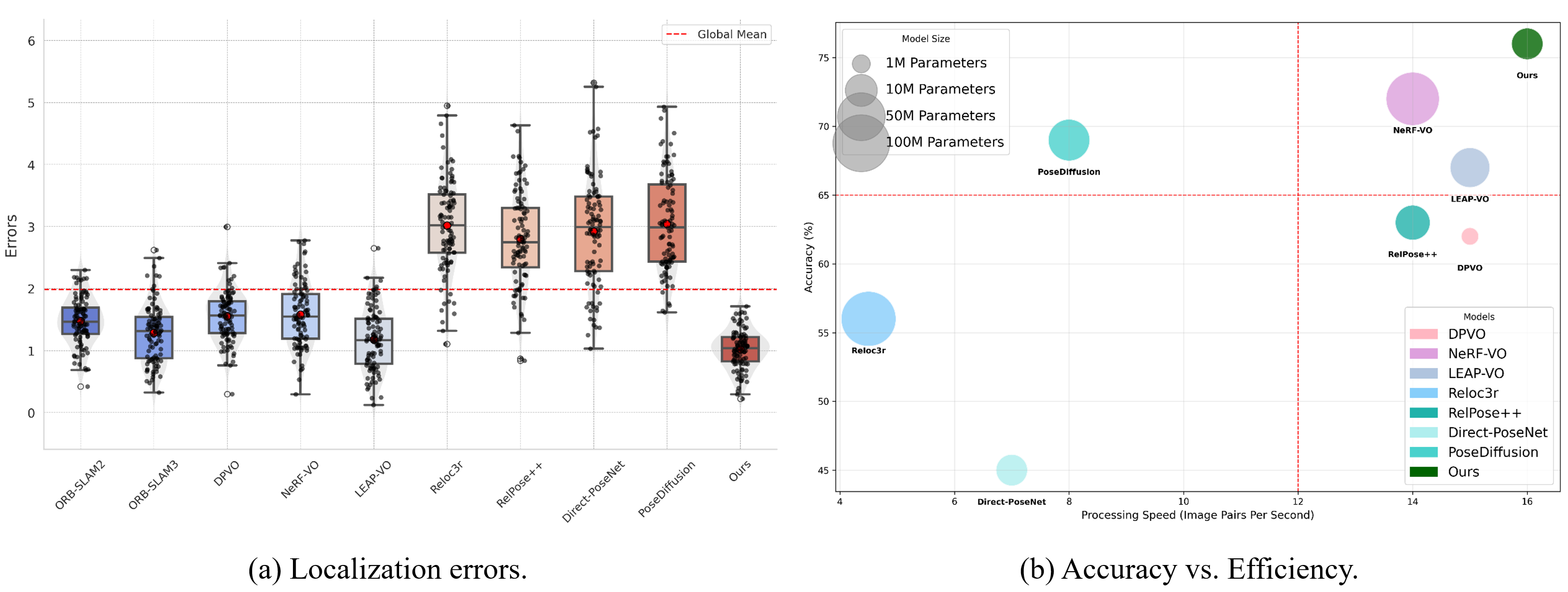}
    \caption{(a) Box plot of the localization errors on the Chang'e-3 landing sequence. (b) Error vs. Efficiency of different models (lower error is better). Bubble size indicates model parameters.}
    \label{fig:boxplot}
\end{figure*}

The box plot in Fig.~\ref{fig:boxplot}(a) further illustrates the superior consistency of KANLoc. The occlusions from lander components severely impact feature-based methods. Our mask augmentation strategy proves effective, enabling the KAN module to produce reliable pose estimates even with partial views. Fig.~\ref{fig:boxplot}(b) visualizes the accuracy-efficiency trade-off, showing that KANLoc occupies a favorable position with high accuracy, high speed, and a moderate model size. Comparisons to structure-based localization (e.g., hierarchical local-feature pipelines) are deferred to future work.

\begin{table}[htp]
    \centering
    \caption{Results of the proposed method and other methods on the real flight dataset. The height error (Err. Height) is explicitly analyzed to address scale ambiguity concerns.}
    \begin{tabular}{lcccc}
    \toprule
    \textbf{Method} & \textbf{Err. Trans.} & \textbf{Err. Rot.} & \textbf{Err. Height} & \textbf{FPS}$\uparrow$ \\\midrule
        \textbf{ORB-SLAM2}$^{\dagger}$ & 17.10 & 15.85 & 11.20 & 18.54 \\
        \textbf{ORB-SLAM3}$^{\dagger}$ & 13.60 & 12.92 & 9.35 & \textbf{19.41} \\
        \textbf{DPVO} & 16.05 & 14.80 & 10.50 & 14.61 \\
        \textbf{NeRF-VO} & 12.90 & 13.25 & 8.10 & 13.80 \\
        \textbf{LEAP-VO} & 12.80 & 12.10 & 7.95 & 14.10 \\\hline
        \textbf{HLoc} & 28.40 & 24.50 & 18.30 & 8.75 \\
        \textbf{Reloc3r} & 33.50 & 28.90 & 22.40 & 4.15 \\
        \textbf{RelPose++} & 29.90 & 25.70 & 19.80 & 7.01 \\
        \textbf{Direct-PoseNet} & 21.40 & 18.50 & 14.20 & 6.26 \\
        \textbf{PoseDiffusion} & 20.70 & 17.80 & 13.50 & 3.89 \\\hline
        \textbf{KANLoc (Ours)} & \textbf{12.50} & \textbf{9.25} & \textbf{5.14} & 15.52 \\   
    \bottomrule
    \end{tabular}
    \label{tab:real_results}
\end{table}

\subsection{Ablation Study}
{We conducted comprehensive ablation studies using the synthetic dataset to ensure a controlled evaluation with precise ground truth.} Specifically, we evaluated the impact of mask augmentation strategies, absolute localization frequency, loss function weights, and the choice of the pose regressor architecture on the overall system performance.

\subsubsection{Mask Augmentation}
\label{sec:mask_ablation}
To simulate occlusions from lander components, we apply random rectangular masks to training images. {We define three augmentation levels based on occlusion ratios: \textit{Low} (5 masks, $\sim$10\%), \textit{Medium} (10 masks, $\sim$20\%), and \textit{High} (20 masks, $\sim$40\%), with mask dimensions sampled from $[0.1H, 0.2H]$. As shown in Table~\ref{tab:mask_ablation}, medium augmentation yields the best performance, balancing feature diversity with information preservation, whereas excessive masking degrades accuracy.}

\begin{table}[htp]
    \centering
    \caption{Impact of mask augmentation on localization accuracy. The "Occlusion Ratio" indicates the approximate percentage of the image area covered by masks.}
    \label{tab:mask_ablation}
    \begin{tabular}{lcccc} 
    \toprule
    \textbf{Level} & \textbf{Count} & \textbf{\makecell{Occlusion\\Ratio}} & \textbf{Trans. (m)$\downarrow$} & \textbf{Rot. (\degree)$\downarrow$} \\ 
    \midrule
    None      & 0   & 0\%            & 9.92            & 8.25              \\
    Low       & 5   & $\sim$10\%     & 7.78            & 7.10              \\
    Medium    & 10  & $\sim$20\%     & \textbf{6.72}   & \textbf{7.02}     \\
    High      & 20  & $\sim$40\%     & 13.85           & 10.18             \\
    \bottomrule
    \end{tabular}
\end{table}

\subsubsection{Absolute Localization Frequency}
We investigated the impact of the absolute pose estimation frequency on accuracy and efficiency by varying the interval at which absolute constraints are added to the BA (every 20, 10, and 5 frames). This analyzes the trade-off between drift correction and real-time performance.

\begin{table}[htp]
    \centering
    \caption{Impact of absolute pose estimation frequency.}
    \begin{tabular}{l|c|c|c|c}
    \toprule
    \textbf{Frequency} & \textbf{Frames} & \textbf{Trans. (m)$\downarrow$} & \textbf{Rot. (\degree)$\downarrow$} & \textbf{FPS}$\uparrow$ \\ \midrule
    Low & 20       & 13.10            & 11.38              & \textbf{19.5} \\
    Medium & 10   & \textbf{6.85}            & \textbf{6.12}              & 16.2         \\
    High & 5      & 4.72   & 5.02    & 6.5         \\
    \bottomrule
    \end{tabular}
    \label{tab:frequency_ablation}
\end{table}

Table~\ref{tab:frequency_ablation} shows that increasing the frequency reduces error but also lowers the processing speed. Low-frequency corrections are fastest but less accurate; high-frequency corrections are most accurate but too slow for real-time use. A medium frequency (every 10 frames) offers the best balance between accuracy and real-time performance.

\subsubsection{Loss Function}
We conducted an ablation study on the loss weight $\lambda$, which balances the translation and rotation terms. We tested $\lambda$ values of 1, 5, 10, and 20.

\begin{table}[htp]
    \centering
    \caption{Effect of different loss weight ratios on localization accuracy. Evaluated on a validation split.}
    \begin{tabular}{c|c|c|c}
    \toprule
    $\lambda$ & \textbf{Translation (m)$\downarrow$} & \textbf{Rotation ($^\circ$)$\downarrow$} & \textbf{Conv. Steps (×1k)} \\
    \midrule
    1   & 10.42 & 5.15 & 320 \\
    5   & 7.18 & 5.58 & 280 \\
    10  & \textbf{6.73} & \textbf{6.61} & \textbf{240} \\
    20  & 3.94 & 7.20 & 170 \\
    \bottomrule
    \end{tabular}
    \label{tab:loss_ablation}
\end{table}

Table~\ref{tab:loss_ablation} shows that increasing $\lambda$ reduces translation error at the cost of higher rotation error. The best balance between accuracy for both terms and convergence speed is achieved at $\lambda=10$, which we use for all experiments.

\subsubsection{Pose Regressor Ablation}
To validate our choice of a KAN-based pose regressor, we replaced it with parameter-matched MLP and Transformer backbones, keeping all other system components and training protocols fixed. As shown in Table~\ref{tab:regressor_ablation_condensed}, the KAN architecture is not only the most accurate with full training data, but also demonstrates superior sample efficiency. When trained on only 25\% of the data, KAN's translation error degrades by just 23\%, compared to 50\% for the Transformer and 65\% for the MLP. This efficiency stems from KAN's learnable spline-based activations, which provide an adaptive basis for approximating the complex image-to-pose mapping more effectively than the fixed nonlinearities of an MLP or the general attention mechanism of a Transformer, especially with limited data. This ablation confirms that the KAN regressor is a key contributor to KANLoc's overall performance.

\begin{table}[htp]
    \centering
    \caption{Ablation of the pose regressor architecture (mean RMSE on a validation split). KAN demonstrates superior accuracy and sample efficiency.}
    \label{tab:regressor_ablation_condensed}
    \begin{tabularx}{\linewidth}{l|c|cc|cc}
    \toprule
    \multirow{2}{*}{\textbf{Regressor}} & \multirow{2}{*}{\textbf{Params}} & \multicolumn{2}{c|}{\textbf{Train Data: 25\%}} & \multicolumn{2}{c}{\textbf{Train Data: 100\%}} \\
    \cmidrule(lr){3-4} \cmidrule(lr){5-6}
    & \textbf{(M)} & \textbf{Trans(m)} & \textbf{Rot($^\circ$)} & \textbf{Trans(m)} & \textbf{Rot($^\circ$)} \\
    \midrule
    KAN         & 11.2 & 8.25 & 7.90 & \textbf{6.72} & \textbf{7.02} \\
    MLP         & 12.0 & 13.10 & 13.40 & 7.95 & 8.10 \\
    Transformer & 11.8 & 10.95 & 11.20 & 7.30 & 7.55 \\
    \bottomrule
    \end{tabularx}
\end{table}

\subsubsection{Remark on Generalization} {While absolute errors vary with terrain complexity (Table~\ref{tab:synthetic_results} and Table~\ref{tab:real_results}), the relative performance trends are expected to generalize across sites. This is because Masking and KANs address intrinsic learning challenges, such as overfitting and high-frequency modeling, independent of specific coordinates.}

\subsection{Deployment on AGX Orin}
{To address concerns regarding the suitability of our approach for onboard deployment and to validate the efficiency of the proposed KAN-based architecture, we conducted hardware-in-the-loop experiments using an NVIDIA Jetson AGX Orin (64GB).  Table~\ref{tab:orin_fps} presents the inference speed (FPS) of KANLoc compared to ORB-SLAM3 and RelPose++ under different power budgets (15W, 30W, 50W, MAXN). KANLoc consistently outperforms both baselines across all power modes, achieving real-time performance even at the lowest power setting. Regarding resource consumption, the model utilizes approximately 8.3 GB of VRAM and 35\% of the RTX Orin's compute capacity during inference, leaving ample margin for parallel tasks. This demonstrates that our method is not only accurate but also efficient enough for practical deployment on resource-constrained embedded systems commonly used in lunar landers.}

\begin{table}[htbp]
    \centering
    \caption{Inference Speed (FPS) Comparison on NVIDIA Jetson AGX Orin under different power budgets.}
    \label{tab:orin_fps}
    \renewcommand{\arraystretch}{1.2}
    \begin{tabular}{l|c|c|c|c}
    \toprule
    \multirow{2}{*}{\textbf{Method}} & \multicolumn{4}{c}{\textbf{Power Mode (Power Budget)}} \\ \cline{2-5} 
    & \textbf{15W} & \textbf{30W} & \textbf{50W} & \textbf{MAXN} \\ \midrule
    ORB-SLAM3 & \textbf{9.85} & \textbf{12.52} & \textbf{15.60} & \textbf{17.90} \\ 
    HLoc & 2.15 & 4.35 & 6.50 & 7.80 \\
    \textbf{KANLoc (Ours)} & 8.40 & 12.10 & 14.20 & 14.85 \\ \bottomrule
    \end{tabular}
\end{table}

\section{Discussion}
\subsection{Generalization and Uncertainty Estimation} 
{It is important to note that the proposed AVL regressor is scene-specific, implicitly encoding the metric map of the target site within its weights. While the method requires site-specific training, the masking strategy effectively bridges the Sim-to-Real domain gap, allowing the model to generalize from synthetic training data to real-world flight imagery with varying illumination and textures.}

{However, a limitation of the current framework is its deterministic nature. The confidence score relies on a fixed covariance assumption, which may not capture the varying reliability of visual features in degraded environments. Ideally, the network should predict a heteroscedastic uncertainty score to replace the fixed covariance in Algorithm~\ref{alg:kan_avl_compact}}:
\begin{equation}
    s_k \!\leftarrow\! \alpha\,\cos(\mathbf{f}_c^i,\mathbf{f}_{m_k}) + (1-\alpha)\,\mathrm{conf}(T_k^{c_i}{_{w_0}}).
\end{equation}
{Integrating such dynamic confidence into the g2o framework would allow the fusion system to adaptively downweight AVL measurements in degraded scenarios, seamlessly falling back to robust relative odometry to ensure trajectory continuity.}

\subsection{Role of Inertial Sensors and System Redundancy}

{While high-grade Inertial Measurement Units (IMUs) are standard on lunar landers and Visual-Inertial Odometry (VIO) is widely used for drift mitigation, we intentionally focus on a \textit{pure vision} solution to enhance strategic system redundancy. In deep space exploration, reliance on a single sensor modality poses risks; therefore, developing a robust, IMU-independent capability ensures navigation continuity even if the IMU fails or saturates during high-vibration phases. Although pure VO typically suffers from unbounded drift, our fusion architecture effectively addresses this by employing the KAN-based AVL as a global correction mechanism, conceptually similar to loop closure, to periodically reset accumulated error. This enables KANLoc to achieve bounded trajectory accuracy solely through visual inputs, providing a critical fail-safe backup for the primary navigation system.}

\section{Conclusion}
In this paper, we proposed KANLoc, a robust and efficient monocular 6-DoF localization framework for lunar landers. Our core innovation is the KAN that provides sparse yet highly accurate absolute pose anchors, which are fused with visual odometry via constrained bundle adjustment to effectively mitigate cumulative drift. Extensive experiments on synthetic and real-world datasets confirm that KANLoc delivers superior accuracy and real-time performance against baselines.
Primary limitations lie in the domain gap and feature degradation under extreme lighting. 
Future work will prioritize four key directions: 
1) Probabilistic Modeling: Extending the KAN regressor to predict uncertainty scores for adaptive sensor fusion; 
2) Hardware Migration: Optimizing the model via quantization for deployment on space-grade embedded platforms (e.g., FPGAs or SoCs); 
3) Environmental Robustness: Incorporating physics-based modeling of dust plumes and outgassing to rigorously test resilience during touchdown; 
and 4) Closed-loop Validation: Integrating the system into Hardware-in-the-loop (HIL) GNC simulations to verify stability in active landing control.

\end{document}